\documentclass[10pt,twocolumn,letterpaper]{article}

\usepackage{3dv}
\makeatletter
\@namedef{ver@everyshi.sty}{}
\makeatother

\usepackage[nolist]{acronym}
\usepackage[symbols]{glossaries-extra}
\usepackage{times}
\usepackage{epsfig}
\usepackage{graphicx}
\usepackage{amsmath}
\usepackage{amssymb}
\usepackage{subcaption}
\usepackage{makecell}
\usepackage{placeins}
\usepackage{float}
\usepackage{adjustbox}

\usepackage{tikz}
\usepackage{pgfplots}
\usetikzlibrary{positioning, shapes, arrows, math}
\tikzset{>=stealth',pil/.style={->, thick, shorten <=2pt, shorten >=2pt,}}

\threedvfinalcopy 

\def\threedvPaperID{108} 

\ifthreedvfinal\fi

\glsxtrnewsymbol[description={color}]{RGB}{\ensuremath{{(R, G, B)}^{\top}}}
\glsxtrnewsymbol[description={point in 3D space}]{XYZ}{\ensuremath{{(X, Y, Z)}^{\top}}}
\glsxtrnewsymbol[description={orientation}]{thetaphi}{\ensuremath{{(\theta, \phi)}^{\top}}}
\glsxtrnewsymbol[description={image coordinate}]{xy}{\ensuremath{{(x, y)}^{\top}}}
\glsxtrnewsymbol[description={view coordinate}]{uv}{\ensuremath{{(u, v)}^{\top}}}

\glsxtrnewsymbol[description={5D light field}]{L5D}{\ensuremath{L(X, Y, Z, \theta, \phi)}}

\glsxtrnewsymbol[description={4D light field}]{L}{\ensuremath{L(x, y, u, v)}}
\glsxtrnewsymbol[description={single view}]{Iuv}{\ensuremath{I_{u, v}}}
\glsxtrnewsymbol[description={center view}]{I00}{\ensuremath{I_{0, 0}}}
\glsxtrnewsymbol[description={horizontal epipolar-plane image}]{Iyv}{\ensuremath{I_{y, v}}}
\glsxtrnewsymbol[description={vertical epipolar-plane image}]{Ixu}{\ensuremath{I_{x, u}}}

\glsxtrnewsymbol[description={baseline}]{b}{\ensuremath{b}}
\glsxtrnewsymbol[description={focal length}]{f}{\ensuremath{f}}
\glsxtrnewsymbol[description={disparity}]{d}{\ensuremath{d}}
\glsxtrnewsymbol[description={depth}]{Z}{\ensuremath{Z}}

\glsxtrnewsymbol[description={pixel}]{s}{\ensuremath{s}}
\glsxtrnewsymbol[description={particle}]{p}{\ensuremath{c}}
\glsxtrnewsymbol[description={size of particle set}]{K}{\ensuremath{K}}
\glsxtrnewsymbol[description={number of PatchMatch iterations}]{N}{\ensuremath{N}}
\glsxtrnewsymbol[description={schedule for iteration $i$}]{phii}{\ensuremath{\phi_{i}}}
\glsxtrnewsymbol[description={particle set for pixel $s$}]{Ps}{\ensuremath{C_s}}
\glsxtrnewsymbol[description={extended particle set for pixel $s$}]{Rs}{\ensuremath{R_s}}
\glsxtrnewsymbol[description={interval of disparity values}]{drange}{\ensuremath{\left(d_{\min}, d_{\max}\right)}}
\glsxtrnewsymbol[description={energy of particle $p$ at pixel $s$}]{Es}{\ensuremath{E_s}}
\glsxtrnewsymbol[description={probability of particles at pixel $s$}]{ps}{\ensuremath{p_s}}
\glsxtrnewsymbol[description={data term}]{psidata}{\ensuremath{\psi_{data}}}
\glsxtrnewsymbol[description={smoothness term}]{psismooth}{\ensuremath{\psi_{smooth}}}
\glsxtrnewsymbol[description={neighborhood of pixel $s$}]{Ns}{\ensuremath{N(s)}}

\glsxtrnewsymbol[description={shifting index for~\ac{EPI}}]{ixu}{\ensuremath{i_{x, u}}}
\glsxtrnewsymbol[description={adjusted shifting index for~\ac{EPI} pixel $(x, u)$}]{iixu}{\ensuremath{i'_{x, u}}}
\glsxtrnewsymbol[description={shifting kernel index for~\ac{EPI}}]{kxu}{\ensuremath{k_{x, u}}}
\glsxtrnewsymbol[description={occlusion mask for~\ac{EPI} column $x$}]{mx}{\ensuremath{m_x}}

\glsxtrnewsymbol[description={L1 loss function}]{L1}{\ensuremath{\mathcal{L}^1}}
\glsxtrnewsymbol[description={L2 loss function}]{L2}{\ensuremath{\mathcal{L}^2}}
\glsxtrnewsymbol[description={Mean Squared Error loss function}]{Lmse}{\ensuremath{\mathcal{L}_{mse}}}
\glsxtrnewsymbol[description={expected loss function}]{LE}{\ensuremath{\mathcal{L}_E}}
\glsxtrnewsymbol[description={one-cold loss function}]{L1c}{\ensuremath{\mathcal{L}_{cold}}}
\glsxtrnewsymbol[description={weight between data and smoothness term}]{alphai}{\ensuremath{\alpha_i}}

\begin{document}
\title{Learning to Think Outside the Box: \\ Wide-Baseline Light Field Depth Estimation with EPI-Shift}

\author{Titus Leistner\textsuperscript{1}, Hendrik Schilling\textsuperscript{1}, Radek Mackowiak\textsuperscript{2}, Stefan Gumhold\textsuperscript{3}, Carsten Rother\textsuperscript{1}\\
Visual Learning Lab, Heidelberg University\textsuperscript{1}, Robert Bosch GmbH\textsuperscript{2}, TU Dresden\textsuperscript{3}\\
{\tt\small first.last@iwr.uni-heidelberg.de\textsuperscript{1}, first.last@de.bosch.com\textsuperscript{2}, first.last@tu-dresden.de\textsuperscript{3}}
}

\maketitle
\thispagestyle{empty}

\begin{acronym}[]
    \acro{EPI}{Epipolar-Plane Image}
    \acro{SPO}{Spinning Parallelogram Operator}
    \acro{RPCA}{Robust Principal Component Analysis}
    \acro{SVD}{Singular Value Decomposition}
    \acro{NP}{Nondeterministic Polynomial Time}
    \acro{MRF}{Markov Random Field}
    \acro{SAD}{Sum of Absolute Gradients}
    \acro{GRAD}{Sum of Gradient Differences}
    \acro{ZNCC}{Zero-mean Normalized Cross Correlation}
    \acro{RANSAC}{Random Sample Consensus}
    \acro{SIFT}{Scale-Invariant Feature Transform}
    \acro{ReLU}{Rectified Linear Unit}
    \acro{BN}{Batch Normalization}
    \acro{VR}{Virtual Reality}
    \acro{MSE}{Mean Squared Error}
    \acro{MAE}{Mean Absolute Error}
    \acro{GPU}{Graphics Processing Unit}
    \acro{CNN}{Convolutional Neural Network}
    \acro{LIDAR}{Light Detection and Ranging}
\end{acronym}

\begin{abstract}
We propose a method for depth estimation from light field data, based on a fully convolutional neural network architecture.
Our goal is to design a pipeline which achieves highly accurate results for small- and wide-baseline light fields.
Since light field training data is scarce, all learning-based approaches use a small receptive field and operate on small disparity ranges.
In order to work with wide-baseline light fields, we introduce the idea of \acs{EPI}-Shift:
To virtually shift the light field stack which enables to retain a small receptive field, independent of the disparity range.
In this way, our approach ``learns to think outside the box of the receptive field''.
Our network performs joint classification of integer disparities and regression of disparity-offsets.
A U-Net component provides excellent long-range smoothing.
\acs{EPI}-Shift considerably outperforms the state-of-the-art learning-based approaches and is on par with hand-crafted methods.
We demonstrate this on a publicly available, synthetic, small-baseline benchmark and on large-baseline real-world recordings.
\end{abstract}

\section{Introduction}
%
\begin{figure}[t]
    \begin{center}
        \begin{subfigure}{\linewidth}
            \includegraphics[width=0.633\linewidth]{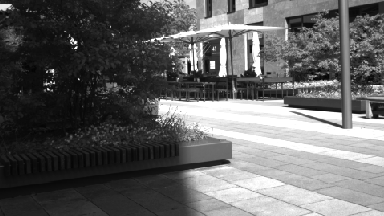}
            \includegraphics[width=0.357\linewidth]{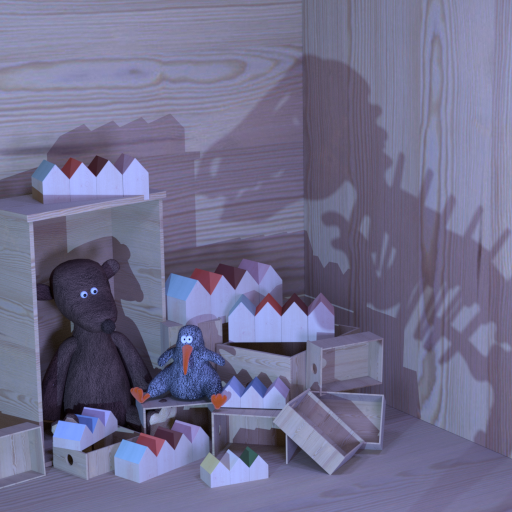}
            \subcaption{\textbf{Center view} of real (left) and synthetic (right) light field\label{fig:teaser_center}\vspace{0.15cm}}
        \end{subfigure}
        \begin{subfigure}{\linewidth}
            \includegraphics[width=0.633\linewidth]{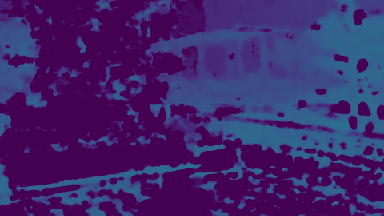}
            \includegraphics[width=0.357\linewidth]{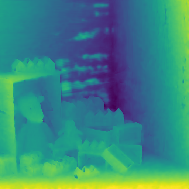}
            \subcaption{\textbf{EPINET-Cross}~\cite{Shin2018}\label{fig:teaser_epinet}\vspace{0.15cm}}
        \end{subfigure}
        \begin{subfigure}{\linewidth}
            \includegraphics[width=0.633\linewidth]{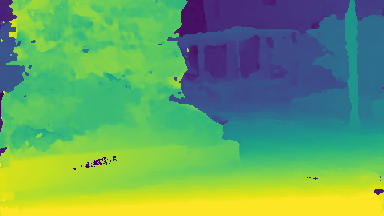}
            \includegraphics[width=0.357\linewidth]{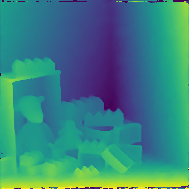}
            \subcaption{Our \textbf{\acs{EPI}-Shift}\label{fig:teaser_epishift}}
        \end{subfigure}
    \end{center}
    \vspace{-0.5cm}
     \caption{
     \textbf{Light Field Depth Estimation.} 
     \textbf{(a)} A {\it real} light field (left) with a large disparity range of $[0, 12]$ and a {\it synthetic} light field (right) with a small disparity range of $[-2, 2]$.
     \textbf{(b)} The current state-of-the-art method EPINET-Cross~\cite{Shin2018} has only been trained for the small disparity range.
     It therefore fails at extreme disparities in the synthetic image (right, background) and outside of the trained range in the real image (left, foreground).
     \textbf{(c)} Our~\acs{EPI}-Shift approach performs well for both, small and large disparities. Due to better generalization, it even outperforms EPINET-Cross for small disparity ranges.
     \label{fig:teaser}
     }
\end{figure}

In the current deep learning era, autonomous driving and intelligent robotics are becoming more and more practical.
For those applications, and many more, accurate depth perception is often a crucial component.
Depth estimation can be performed with a wide variety of different hardware setups.
On one end of the spectrum there is monocular depth estimation from single cameras~\cite{Fu2018}\cite{Guo2018}.
In between, there is stereo-matching from rectified images captured by two cameras~\cite{Tulyakov2018}\cite{Sunghoon2019}.
At the other end of the spectrum there is depth estimation from light field camera arrays~\cite{Wilburn2001} which is the focus of this work. 
Light field depth estimation can be seen as a well-structured multi-view stereo approach.
Although the hardware setup of a light field is more involved than a stereo setup, the depth information ``encoded'' in light fields is considerably more precise and inherently less sensitive to occlusions, due to the multiple redundant viewpoints. 

\captionsetup[figure]{skip=12pt}
\begin{figure}[t]
    \includegraphics[width=\linewidth]{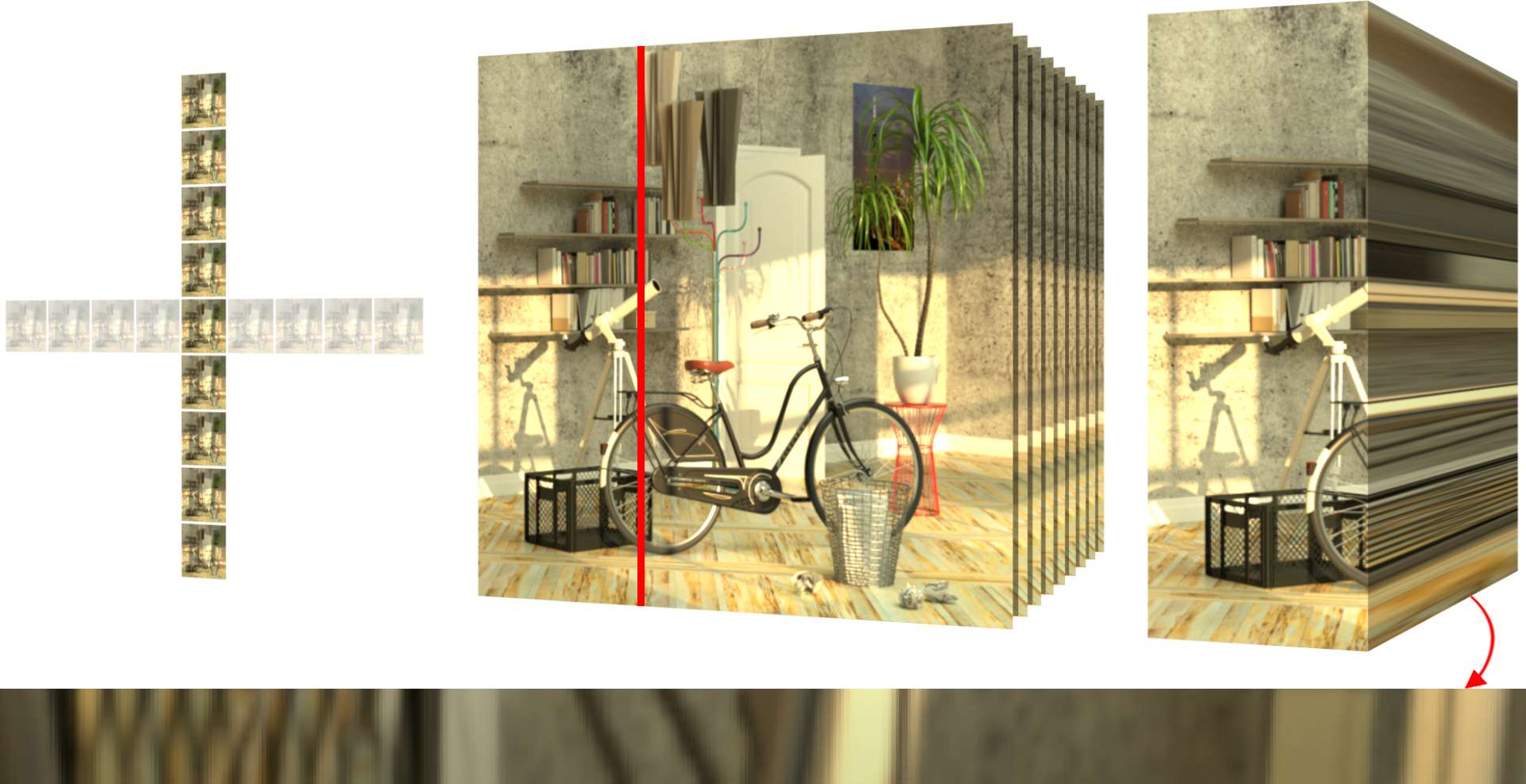}
    \caption{
    \textbf{Extraction of an \acf{EPI}}. The vertical views captured by a \textbf{cross light field} (left) are stacked (middle).
    A slice through a column (right) forms an \textbf{\ac{EPI}} (bottom).
    Note, how the slopes of lines in the~\ac{EPI} encode the three-dimensional structure of the scene.
    \label{fig:epi_extraction}
    \vspace{-0.3cm} 
    }
\end{figure}


There are two types of light field cameras. 
First, a multi-camera setup, see \eg a cross-camera in Figure~\ref{fig:epi_extraction} with its corresponding~\ac{EPI}.
Such a setup is unfortunately expensive, difficult to assemble, synchronize and calibrate, compared to~\eg a stereo setup.
However, once these challenges are mastered, the major advantage of a multi-camera system is its accuracy due to the wide baseline, since reconstruction accuracy grows linearly with the baseline between viewpoints and hence, a small baseline yields inferior accuracy.
We therefore focus on this setup.
Second, a plenoptic camera based on a micro-lens array~\cite{Ng2005} which has however, a rather limited resolution and baseline.

In this work, we propose a learning-based light field depth estimation method.
This is challenging due to the non-availability of real-world training data.
The creation of real-world reference depth is problematic, as no other dense measurement approach is more accurate than light field depth estimation.
For example, structured light scanning is problematic in the context of occlusions and~\ac{LIDAR} is considerably more sparse than light field data.
Therefore, all training data is synthetic and the pool of publicly available datasets is small.
Also, all these training images have a small baseline, emulating a micro-lens based camera rather than a camera array.
Hence, current learning-based approaches fail dramatically for wide-baseline light fields, as exemplified in the real world scene in Figure~\ref{fig:teaser}.
Interestingly, even for smaller disparities, the performance is limited by poor generalization, as demonstrated in the synthetic scene in Figure~\ref{fig:teaser}, where artifacts appear {\it within} the trained disparity range.

\captionsetup[figure]{skip=0pt}
\begin{figure}[t]
    \includegraphics[width=\linewidth]{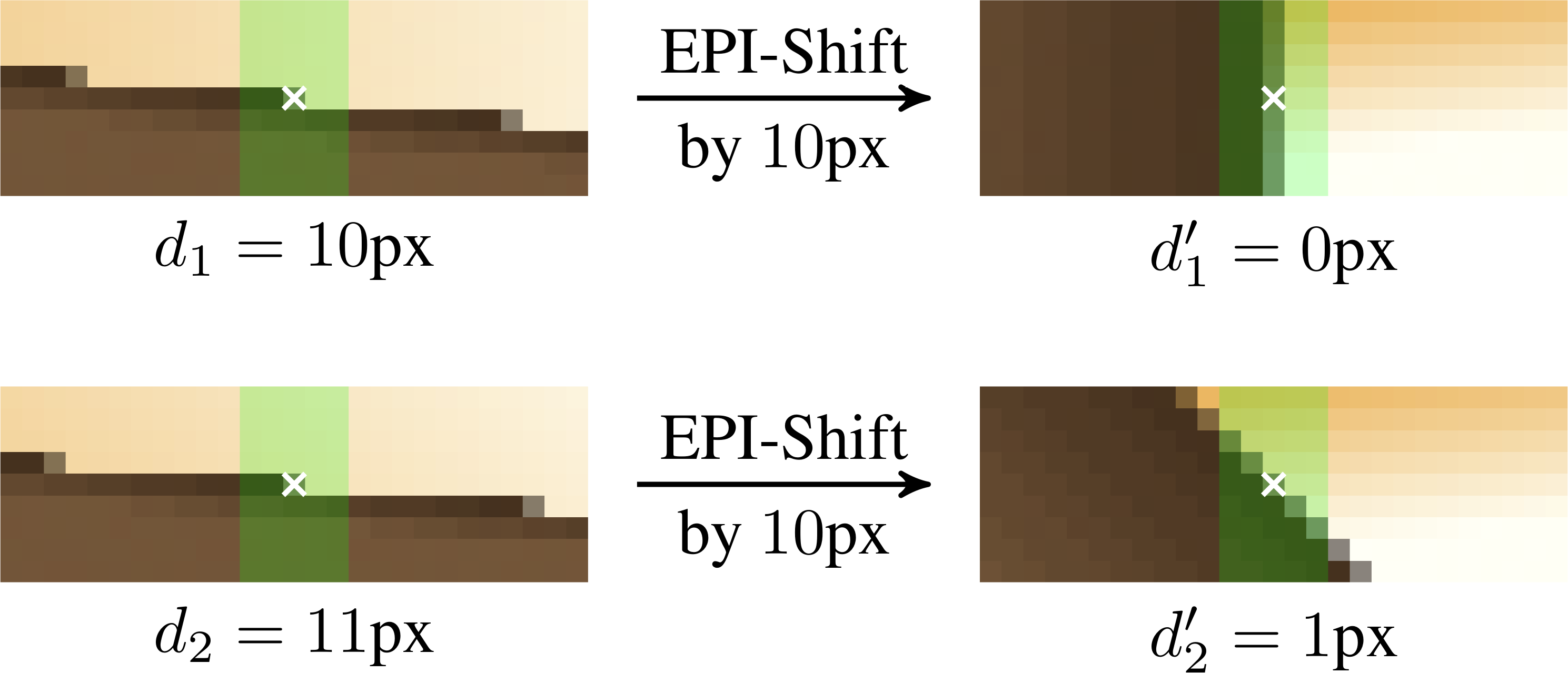}
    \vspace{-0.2cm} 
    \caption{
        \textbf{Idea of \ac{EPI}-Shift.}
        Two~\textbf{\acp{EPI}} (left), consisting of horizontal lines from different cameras.
        The~\acsp{CNN} task is to estimate the correct disparity for the center pixel (white cross) by predicting the slope of the line going through this pixel.
        Note the nearly invisible difference between a disparity of $10$ and $11$ pixels which can only be estimated using a large receptive field.
        After applying an \textbf{EPI-Shift} of $10$ pixels (right), the difference is clearly visible.
        Therefore, our network only requires a minimal receptive field (green box) to classify whether the shifted disparity lies within $\pm0.5$ pixels and regress a sub-pixel accurate disparity offset.
        \label{fig:basic_idea}
    }
\end{figure}
%
%

One major cause for this problem, the limited receptive field of previous methods, is illustrated in Figure \ref{fig:basic_idea}.
Expanding the receptive field would cause worse generalization performance.
However, by applying~\ac{EPI}-Shifts to the input of our neural network, we circumvent this flaw.
The basic idea is inspired by the technique of plane-sweep volumes.
Instead of directly estimating the disparity from the light field image stacks, we utilize a plane, sweeping through space, as common for stereo and multi-view depth estimation~\cite{Collins1996}\cite{Sunghoon2019}.
Hence, we split the task into classification and regression.

The classification map states, per pixel, whether objects observed at tested plane sweep are within a refocused disparity range of $\pm0.5$.
It is used to merge all independent estimates from the plane sweep volume, while the disparity regression provides sub-pixel accuracy.
This approach considerably improves generalization, as we are now able to infer the depth of a wide-baseline scene with a network, trained solely with small-baseline training data.

Let us summarize our main contributions:
\begin{itemize}
\item Applying the idea of plane-sweep volumes in the context of light fields, which we denote as \ac{EPI}-Shift. This enables learning-based approaches to generalize well to large-disparity test data, even with small-disparity training data.
\item A network architecture, which enables improved long-range reasoning by combining a feature extraction network~\cite{Shin2018} with a subsequent U-Net architecture~\cite{Ronneberger2015}\cite{Heber201602} for excellent long-range smoothing with low artifacts.
\item Our approach considerably outperforms the state-of-the-art learning-based approaches with same input modality and is on par with hand-crafted methods.
\end{itemize}

\section{Related Work}
Light field depth estimation methods can be categorized into optimization-based and learning-based approaches.

\subsection{Methods based on Optimization}

Heber and Pock~\cite{Heber2014} applied~\ac{RPCA} to light field depth estimation.
Their method measures the quality of image alignments by projecting images to column vectors of a shared matrix, followed by a convex optimization of stereo matching and denoising.
However, the method is vulnerable to occlusions.


Jeon \etal~\cite{Jeon2015} specifically address lenslet cameras, proposing a novel distortion correction.
The actual depth estimation is performed using a pixel-wise cost volume inspired by traditional stereo matching techniques which is combined with discrete Fourier transform.
To ensure smoothness, the algorithm applies a sparse matching using graph cut at salient feature points based on~\ac{SIFT}.
The method shows a good performance on real images from micro-lense cameras but suffers from noise and artifacts at occlusion boundaries.


Lin \etal~\cite{Lin2015} utilize a focal stack composed of light field data for depth estimation.
Their method is based on the local symmetry of the color distribution in a patch around the real depth.
Smoothness is ensured by an optimization with graph cut.
Unfortunately, the refocusing of micro lense images produces slight artifacts, making the approach not as accurate as most~\ac{EPI}-based methods.

Wanner and Goldluecke~\cite{Wanner2014} proposed a solution that calculates the disparity directly, using linear algebra operations.
To estimate the line slope, an adapted structure tensor is applied to the 4D light field.
The strongest eigenvector of this tensor, composed of all partial derivations, is aligned with the~\ac{EPI} gradient.
As this method produces inaccuracies in non-textured regions and outliers at occlusions, additional optimization methods are required.

Neri \etal~\cite{Neri2015} proposed another method based on $\mathcal{L}^2$ matching costs. 
In order to ensure smoothness, a local optimization of discrete depth labels on a resolution pyramid is performed.
The method still produces artifacts in large non-textured areas and at object boundaries.


Zhang \etal~\cite{Zhang2016} introduced the~\ac{SPO} fitting lines with different slopes through the center pixel of a parallelogram.
Each line divides the parallelogram in two distinct areas.
The~\ac{SPO} computes $\chi^2$ differences of color distributions between those areas, used as confidence measure.
However, the quality at occlusions still depends on the relative angle between~\ac{EPI} extractions and objects boundaries.

Sheng \etal~\cite{Sheng2017} introduced a method to compensate for this effect, utilizing a bigger set of~\acp{EPI} extracted at arbitrary angles.
For occlusion reasoning, the algorithm utilizes the variance of depth estimates at different \ac{EPI} orientations.
Using the direction of the derived occlusion boundary, the optimal non-occluded~\ac{EPI} is selected to determine the final depth value.
The authors also utilize an operator similar to~\ac{SPO} but comparing the color distributions between semicircles instead of parallelograms.
While this method handles occlusions very well, it depends on a large number of views that are not present in all real applications.

In contrast, Schilling \etal~\cite{Schilling2018} introduced a method that does only depend on a cross input.
It performs local optimization based on PatchMatch~\cite{Barnes2009}.
This model produces state-of-the-art results on the HCI Light Field Benchmark~\cite{Honauer2016}.
However, it contains many hand tuned parameters which could probably be improved using a learning-based method.

\subsection{Methods based on Deep Learning}
In 2016 Heber \etal~\cite{Heber2016} introduced the first deep learning based methods for the task, predicting 2D per-pixel hyper plane orientations.
The utilized~\acs{CNN} processes a cross subset of the light field.
A disparity map is then inferred from the hyperplane parameters by optimizing a convex energy functional.
Due to the lack of training scenes, the authors also contributed a randomly generated synthetic dataset.
The approach is a first step in the direction of learned depth from light fields but the results suffer from strong artifacts and blur.

In order to overcome those downsides, the same authors improved their work~\cite{Heber201602}\cite{Heber2017} by utilizing a U-Net~\cite{Ronneberger2015}.
The first paper~\cite{Heber201602} shows visual and quantitative improvement, but suffers from streaking artifacts, adressed in~\cite{Heber2017}.

One of the most recent works by Shin \etal~\cite{Shin2018} deals with light field depth estimation using a fully-convolutional architecture.
The authors introduce a multi-stream network comprised of multiple inputs for the horizontal, vertical and both diagonal light field stacks.
All outputs of the individual streams are then being concatenated and fed into a second~\acs{CNN}.
The approach reaches state-of-the-art results on the HCI 4D Light Field Benchmark~\cite{Honauer2016}.
However, it is limited to the disparity interval seen during training.



\subsection{Plane Sweep Volumes}
The concept of plane sweep volumes was introduced by Collins~\cite{Collins1996} in 1996.
He addresses the problem of multi-image matching. 
Therefore, features from each input image are projected to a set of parallel planes sweeping through space at increasing depth.

Sunghoon \etal~\cite{Sunghoon2019} apply this principle to multi-view stereo, using a~\acs{CNN} architecture.
A cost volume is created by shifting features, extracted from the multiple input images using an encoder-architecture.
Subsequently, the disparity is regressed by aggregating those costs.
This method enables state-of-the-art occlusion handling results for depth reconstruction from stereo recordings.


    
    
        
    
\begin{figure*}[t]
    \begin{center}
        \input{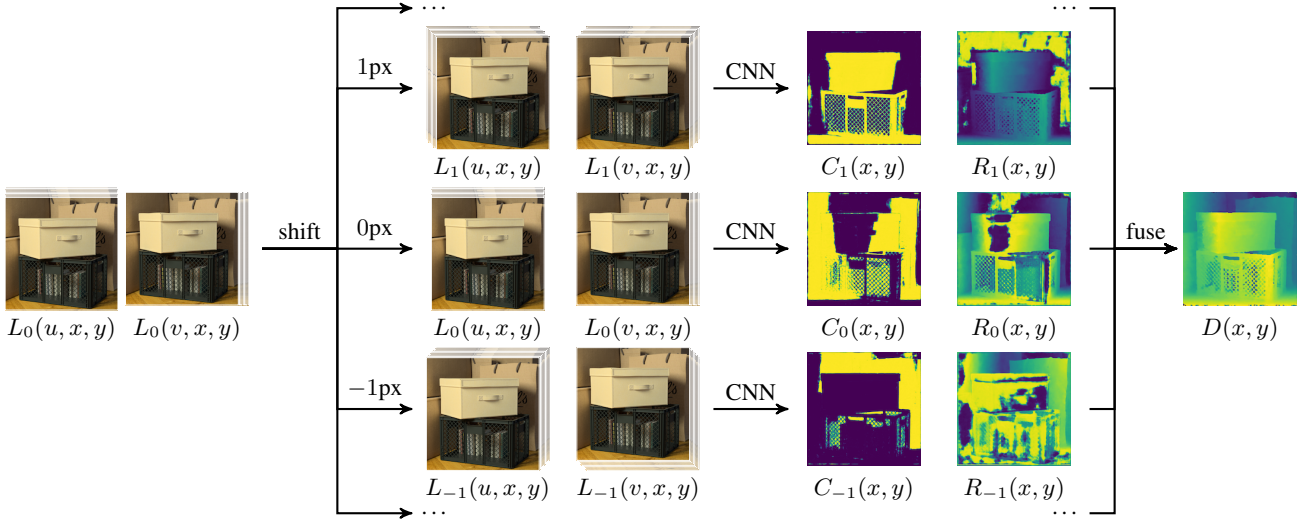}
    \end{center}
    \caption{
        \textbf{Method Overview}.
        The \textbf{input} (left) consisting of two view stacks is shifted several times, producing stacks with different disparity ranges.
        Our \textbf{\ac{CNN}} (middle) processes the shifted stacks, inferring a classification and regression output.
        Each pixel of the \textbf{final result} (right) is assigned to a discrete disparity (classification) and refined by a sub-pixel disparity offset (regression).
        \label{fig:fusion}
    }
\end{figure*}
\FloatBarrier
\section{Method}
The core idea of our method is to use a plane sweep volume~\cite{Collins1996}\cite{Sunghoon2019} and successively apply the same neural network to each depth plane of that volume individually.
The output of the network for each of these disparity ranges are two 2D maps, one for the classification (correct plane/incorrect plane) and one for the disparity offset from the plane, in the range $-0.5$ and $0.5$.
To generate the full disparity map of the scene, for each pixel, the shift with maximum classification activation is chosen to determine the correct plane.
The corresponding per-pixel disparity offset is added to achieve sub-pixel accuracy.
Because we are using a cross light field setup, the plane sweep volume can be constructed using the \ac{EPI}-Shift approach, which refocuses the image stack by applying a skew transformation.

\subsection{Light Field Camera Setup}
\label{method:setup}
Goal of our method is an accurate per-pixel disparity reconstruction within the center view of a $9+8$ cross-shaped light field camera setup with a large baseline.
We limit ourselves to this setup due to the versatile usability in real world scenarios, compared to a star, or full 4D light field setup. 
Many recent submissions to the HCI 4D Light Field Benchmark~\cite{Honauer2016} demonstrate that research is shifting towards using more views from the available $81$ input views, \eg by synthesizing a focal stack from all views. 
However, we argue that this is a symptom of ``benchmark optimization'', because adding more views gives diminishing returns and using less views is more practical in the real world. Note that the best approach~\cite{Schilling2018} is not learning-based and requires only $17$ views, compared to the inferior  EPINET-Star~\cite{Shin2018} setup which requires nearly double the amount of views.


4D light fields are recorded by an array of cameras, arranged on a regular 2D grid, indexed by $(u, v)$.
The baseline represents the distance between two adjacent cameras.
Each camera captures a 2D image with pixels, indexed by $(x, y)$.
Similar to the effect of alternatively closing the left and right eye, objects seem to move between different viewpoints.
A change of viewpoints on the $u$-axis for example, causes movement of a projected object point along the $x$-axis.
The straight lines in image space, which represent this depth dependent movement, are called epipolar lines.
For a cross light field the 2D slices of the light field along the $xu$- and $yv$-planes represent the~\acfp{EPI} (see Figure \ref{fig:epi_extraction})~\cite{Bolles1987}.
The $x$- or $y$-distance between the same object point in two adjacent views is the disparity $d$, measured in pixels and being inversely proportional to the depth.
During camera calibration and rectification, the images often get pre-shifted.
Therefore, also negative disparity values occur in some light field datasets.


\begin{figure*}[t]
    \begin{center}
        \input{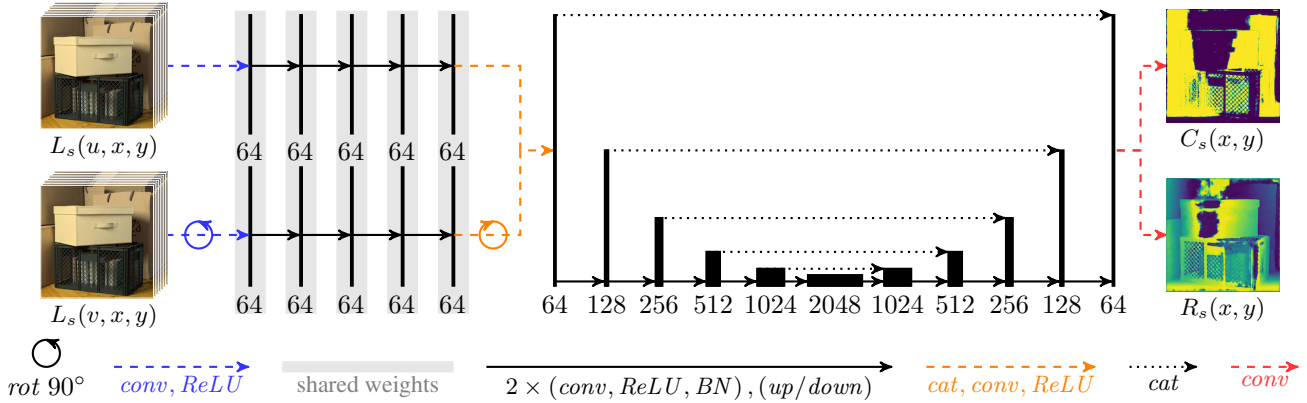}
    \end{center}
    \caption{
    The~\textbf{neural network architecture} of our model, consisting of two parts.
    First, a \textbf{siamese feature extractor}~\cite{Shin2018} (left), with four convolutional blocks for the discovery of local disparity information.
    Second, a \textbf{U-Net} architecture~\cite{Ronneberger2015}\cite{Heber201602} (right) to integrate global information.
    The input consists of two view stacks. 
    Our network outputs a classification of discrete depth labels and a sub-pixel accurate disparity regression.
    The network solely uses convolutional blocks, consisting of two consecutive $3\times3$ convolutions with stride and padding of one.
    Numbers refer to the number of channels.
    }
    \label{fig:architecture}
\end{figure*}
\subsection{EPI-Shift}
Our~\ac{EPI}-Shift approach, which generates the plane sweep volume, boils down to a skew transformation on the $xu$-plane.
Given a 3D stack of views,
\begin{equation}
    L_0(u, x, y)
\end{equation}
defines the color value at a given image position $(x, y)$, recorded by a camera $u$.
The central camera is defined to be at $u=0$.
Positive $u$-indices are assigned to cameras that are located to the right of the central camera.
As visualized in Figure~\ref{fig:basic_idea}, we perform the~\ac{EPI}-Shift by a certain disparity $s$ with
\begin{equation}
    L_s(u, x, y) = L_0(u, x-us, y).
\end{equation}
Note that this operation refers to horizontal~\acp{EPI} only.
However, vertical~\acp{EPI} $L_0(v, x, y)$ behave analogously after a rotation about $90^\circ$ around the $v$-axis.

Because $s$ is defined to be an integer number and we use a cross-shaped setup, no interpolation is required.
We perform nearest-neighbor padding by clipping $x-us$ to the valid pixel range.
However, in order to not waste capacity of the neural network for learning to deal with image-borders, we do not apply any loss in areas effected by the padding.

To enable wide-baseline light field depth estimation, we perform three basic steps, illustrated in Figure~\ref{fig:fusion}.
First, we generate a plane sweep volume by applying the~\ac{EPI}-Shift to the input light field, once per integer disparity within the disparity range of the scene.
Each of those shifts can be thought of as a discrete disparity label.
A pixel is assigned to a certain label if the disparity lies between $-0.5$ and $0.5$, when shifted by the labels disparity.
Second, we infer a classification and a regression map for each of the shifts, using the~\ac{CNN} architecture described in Section~\ref{sec:architecture}.
Third, we compute the final result $D$ for each pixel $(x, y)$ by assigning an integer disparity label
\begin{equation}
    l(x, y)
    =
    \mathit{argmax}_s \left(C_s(x, y)\right)
\end{equation}
according to the shift $s$ that produced the highest classification output $C_s$.
Using the regression map $R_s$ of the respective shift, we add fine-grained disparity information to achieve the sub-pixel accurate result
\begin{equation}
    D(x, y)
    =
    l(x, y) + R_{l(x, y)}(x, y).
\end{equation}

\subsection{Network Architecture}\label{sec:architecture}
Our architecture consists of two parts visualized in Figure~\ref{fig:architecture}.
First, a siamese feature extraction network similar to~\cite{Shin2018}. 
The purpose of this subnetwork is the extraction of local disparity information.
Second, a U-Net architecture (compare~\cite{Ronneberger2015} and~\cite{Heber201602}) with two outputs:
A classification output, assigning discrete per-pixel disparity labels and a continuous regression output, representing the sub-pixel accurate disparity relative to the label.

\paragraph{Siamese Feature Extraction Network:}
The cross light field introduced in Figure~\ref{fig:epi_extraction} provides a horizontal and a vertical stack of input views.
Instead of concatenating the two, we chose a siamese twin architecture, consisting of one subnetwork for each stack.
As both stacks contain similarly aligned~\acp{EPI} after rotation of one stack by $90^\circ$, we share weights between the two subnetworks.
This reduces the number of network parameters and therefore improves generalization.

The feature extraction network contains four fully-convolutional blocks, each consisting of two $3\times3$ convolutions followed by a $\mathit{ReLU}$ activation function and a batch normalization layer each.
We chose a number of $64$ channels and preserve the input dimensions with a padding and stride of one.

To facilitate the classification for the downstream U-Net, we provide it with additional data.
We apply the feature extraction network to both adjacent~\ac{EPI}-Shifts $L_{\pm1}$. 
The extracted features are concatenated with those, extracted from $L_0$ as well as the color information of the center view $L_0(0, x, y)$.
Hence, the number of input channels for the U-Net is, for each shift: The number of channels from the feature extraction subnetwork times two (horizontal + vertical) times three for the three shifts ($-1,0,+1$) plus the center view, i.e. a total of $64 \cdot 2 \cdot 3 + 3 = 387$ channels.
Our experiments showed that these additional inputs improved the distinction between foreground and background objects in ambiguous regions.
This is probably due to the depth hints from adjacent shifts that provide additional information about occlusions and therefore simplify classification.
The addition of the center view allows the joint model to focus on feature extraction in the first part of the network, but still use the center view to guide smoothing in the U-Net part.

\paragraph{U-Net:}
A U-Net~\cite{Ronneberger2015} architecture expands the effective receptive field of the joint model without loss of generalization capability.
It therefore significantly improves the smoothness of non-textured areas.
We chose a depth of five down- and up-sampling layers leading to a receptive radius of $124$ pixels for the U-Net part and $135$ pixels for the whole network.
The concatenated output of the upstream feature extractor network is reduced from $387$ to $64$ channels by an additional $3\times3$ convolution.
For the processing inside the U-Net, we chose the same convolutional blocks as for the feature network, followed by an additional $3\times3$ up- or down-sampling convolution.
The downsampling layers bisect the image dimensions while doubling the number of channels.
Prior to upsampling, the output of the corresponding downsampling is concatenated.
Therefore, the upsampling process doubles the image dimensions but divides the number of channels by four, please see~\cite{Ronneberger2015} for more details.
A final $3\times3$ convolutional layer transforms the $64$ output channels of the U-Net to two channels for the classification and regression output.
Because the regressed disparity can be negative, no final $\mathit{ReLU}$ activation function is applied.


\subsection{Loss Function}
Due to the drastically different outputs of our network, the choice of a well performing loss function is not trivial.
For the classification output, a slight overlap between adjacent shifts might not have an effect on the final result at all.
A large output at distant disparity regions however may cause a misclassification and therefore can destroy the end result.
Our classification loss therefore specifically penalizes those cases.
We define the loss with $C_s(x, y)$ being the classification output for a given shift $s$ at pixel $(x, y)$ as
\begin{equation}
    \mathcal{L}_\mathit{class} 
    = 
    \sum\limits_{s, x, y}
    \left(C_s(x, y) - C^*_s(x, y)\right)^2
    \cdot
    \mathcal{W}_\mathit{disp}(x, y)
\end{equation}
with a disparity weighting of
\begin{equation}
    \mathcal{W}_\mathit{disp}(x, y)
    =
    \left(D(x, y) - D^*(x, y)\right)^2
\end{equation}
computed using the final disparity output $D$ and the ground truth disparity $D^*$ that penalizes misclassifications during training.
The classification ground truth $C^*$ should be high for all pixels within a disparity of $\pm0.5$ pixels.
We therefore tried two different definitions:
First, the one-hot or rectangle function
\begin{equation}
    C^*_s(x, y)
    =
    \begin{cases}
        1  & \quad \text{if } \left|D^*(x, y) - s\right| \leq 0.5 + \epsilon \\
        0  & \quad \text{otherwise}
    \end{cases}
    \label{eq:rect}
\end{equation}
producing hard boundaries between two labels.
Second, the triangle function
\begin{equation}
    C^*_s(x, y)
    =
    \max\left(0.5 + \epsilon - \left|D^*(x, y) - s\right|, 0 \right)
    \label{eq:tri}
\end{equation}
which is more closely related to the regression output.
It therefore should accelerate training and engage the network to share weight capacities between the two.
On the other hand, it outputs lower values at boundaries, which are more vulnerable to misclassifications.
In both cases, we choose a small $\epsilon$ that produces a slight overlap at the border regions between two disparity labels in order to prevent wrong classifications.
We will see in the experiments that the rectangle function (Equation~\ref{eq:rect}) performs slightly better.

The regression output requires smooth surfaces but sharp edges.
We see in our experiments that the $\mathcal{L}^1$ loss function fulfills those requirements for the regression loss best.
We therefore define it as
\begin{equation}
    \mathcal{L}_\mathit{reg} 
    = 
    \sum\limits_{s, x, y}
    \left|R_s(x, y) - D^*(x, y) + s \right|C_s^*(x, y)
    \label{eq:reg}
\end{equation}
with $R$ being the regression output.
We mask out all pixels outside the sub-pixel interval by weighting with the rectangle function $C_s^*$ defined in Equation~\ref{eq:rect}.
In order to compensate for misclassifications at the boundaries of disparity labels, we choose a slightly higher $\epsilon$ than for the classification ground truth.
Due to the fundamentally different trend of the losses during training, a weighting between the two is also  important for computation of the overall loss
\begin{equation}
    \mathcal{L}
    =
    \alpha
    \cdot
    \mathcal{L}_\mathit{reg} 
    +
    \mathcal{L}_\mathit{class}
    .
\end{equation}{}
The choice of a scaling factor $\alpha$ depends on various factors such as the disparity distribution of the training data.

\subsection{Training}
We trained our model on $16$ scenes of the HCI 4D Light Field Benchmark~\cite{Honauer2016} that are not part of the benchmark evaluation.
We implemented the model in PyTorch~\cite{Paszke2017} and trained it for four days on three NVIDIA TITAN X GPUs.
As optimizer we chose Adam with a learning rate of $10^{-4}$ for the first $10000$ iterations.
For another $30000$ iterations, we decreased the learning rate to $10^{-5}$ and fixed the learned batch normalization parameters.

We apply a large variety of data augmentation, comparable to~\cite{Shin2018}, including random color channel re-distribution, random brightness and contrast adjustments, random rotations by multiples of $90^\circ$, random scales between $0.5$ and $1$ and random crops to a patch size of $225\times225$.
This patch size leverages the utilization of global information by the U-Net.
Our training batches contain seven shifts of two stacks extracted from a single RGB light field ($7 \cdot 2 \cdot 3 = 42$ channels).

\subsection{Refinement}
When choosing the rectangle classification ground truth (Equation~\ref{eq:rect}) an additional refinement step can be performed.
In case
\begin{equation}
    \max\limits_s \left(C_s(x, y)\right) < t,
\end{equation}
meaning that no classification exceeds some small threshold $t = 0.01$, we assume that the chosen disparity label at $(x, y)$ would probably be wrong.
In order to smoothly fill this pixel, we apply a median filter to each classification output first.

\section{Experiments}



In this section we present the results of our evaluations.

\subsection{Ablation Studies}
\pgfplotsset{ticks=none}
\pgfplotsset{samples=200}
\begin{table}
    \begin{center}
        \begin{small}
        \begin{tabular}{|l|c|c|c|c|}
        \hline
        Class GT & \makecell{$\epsilon=0.0$ \\ $\alpha=0.25$}  & \makecell{$\epsilon=0.0$ \\ $\alpha=2.5$}  & \makecell{$\epsilon=0.25$ \\ $\alpha=0.25$}  & \makecell{$\epsilon=0.25$\\$\alpha=2.5$} \\
        \hline\hline
            \begin{adjustbox}{margin= 0cm 0cm 0cm 0.1cm}
            \begin{tikzpicture}
            \begin{axis}[
                width=3.1cm, 
                height=2.5cm,
                 xmin = -1.15,
                 xmax = 1.15,
                 ymin = 0,
                 ymax = 1.5, 
                 axis lines = middle,
                 enlargelimits = true
            ]
            \addplot[color=red][domain=-0.5:0.5]coordinates { (-100, 0) (-0.5, 0) (-0.5, 1) (0.5, 1) (0.5, 0) (100, 0)};
            \end{axis}
            \end{tikzpicture}
            \end{adjustbox}
            & $37.25$ & $4.86$ & $5.74$ & $15.47$\\
        \hline
            \begin{adjustbox}{margin= 0cm 0cm 0cm 0.1cm}
            \begin{tikzpicture}
            \begin{axis}[
                width=3.1cm, 
                height=2.5cm,
                 xmin = -1.15,
                 xmax = 1.15,
                 ymin = 0,
                 ymax = 0.75, 
                 axis lines = middle,
                 enlargelimits = true
            ]
            \addplot[color=red]{max(0.5-abs(x), 0)};
            \end{axis}
            \end{tikzpicture}
            \end{adjustbox}
            & $37.43$ & $52.22$ & $24.40$ & $5.27$\\
        \hline
        \end{tabular}
        \end{small}
    \end{center}
    \caption{\acf{MSE} score for the network, trained with \textbf{different classification} ground truth functions $C_s^*$ and values for $\epsilon$ and $\alpha$.}
    \label{tab:ablation}
\end{table}
Because the choice of a classification ground truth function is highly important for our method, we evaluated different functions and parameters.
Table~\ref{tab:ablation} shows the~\ac{MSE} score of our network, trained on either the rectangle function in Equation~\ref{eq:rect} or the triangle function in Equation~\ref{eq:tri}.
As expected, the results show that the triangle function requires a higher $\epsilon$ to compensate for wrong classifications at boundaries of disparity labels.
Due to the slightly better performance of the rectangle function, we chose it for our subsequent evaluations, setting $\epsilon=0.17$ and $\alpha=2.5$.

We also evaluated our~\ac{CNN} architecture, without~\ac{EPI}-Shift, similar to~\cite{Shin2018}.
This model only reached an~\ac{MSE} score of $31.15$ compared to $0.85$ for our model with~\ac{EPI}-Shift.
This clearly indicates that our U-Net architecture requires the shifted~\acp{EPI} in order to properly generalize. 


\subsection{Results on the HCI 4D Light Field Benchmark}
\begin{figure*}[t]
    \centering
    \begin{subfigure}{0.162\textwidth}
        \includegraphics[width=\textwidth]{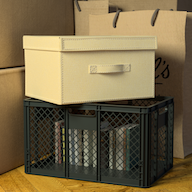}
        \includegraphics[width=\textwidth]{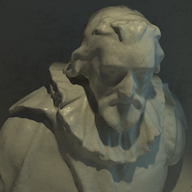}
        \includegraphics[width=\textwidth]{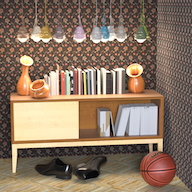}
        \includegraphics[width=\textwidth]{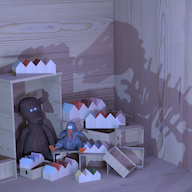}
        \subcaption{Center View}\label{fig:center}
    \end{subfigure}
    \begin{subfigure}{0.162\textwidth}
        \includegraphics[width=\textwidth]{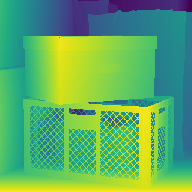}
        \includegraphics[width=\textwidth]{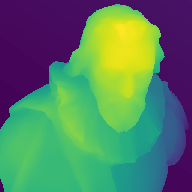}
        \includegraphics[width=\textwidth]{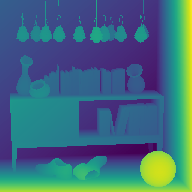}
        \includegraphics[width=\textwidth]{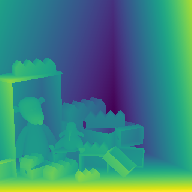}
        \subcaption{Ground Truth}\label{fig:gt}
    \end{subfigure}
    \begin{subfigure}{0.162\textwidth}
        \includegraphics[width=\textwidth]{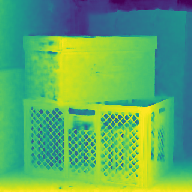}
        \includegraphics[width=\textwidth]{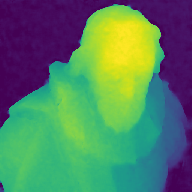}
        \includegraphics[width=\textwidth]{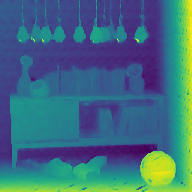}
        \includegraphics[width=\textwidth]{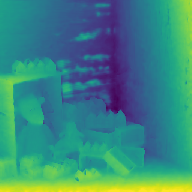}
        \subcaption{EPINET}\label{fig:epinet}
    \end{subfigure}
    \begin{subfigure}{0.162\textwidth}
        \includegraphics[width=\textwidth]{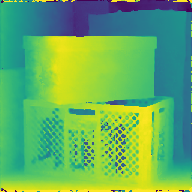}
        \includegraphics[width=\textwidth]{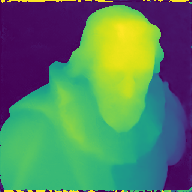}
        \includegraphics[width=\textwidth]{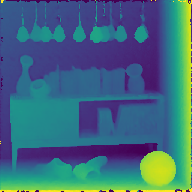}
        \includegraphics[width=\textwidth]{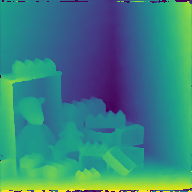}
        \subcaption{Ours}\label{fig:ours}
    \end{subfigure}
    \begin{subfigure}{0.162\textwidth}
        \includegraphics[width=\textwidth]{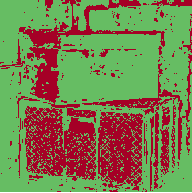}
        \includegraphics[width=\textwidth]{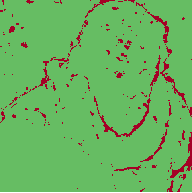}
        \includegraphics[width=\textwidth]{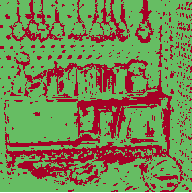}
        \includegraphics[width=\textwidth]{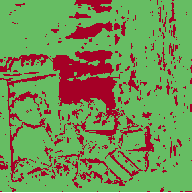}
        \subcaption{EPINET BadPix}\label{fig:epinet_badpix}
    \end{subfigure}
    \begin{subfigure}{0.162\textwidth}
        \includegraphics[width=\textwidth]{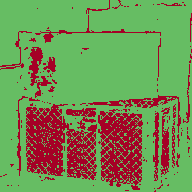}
        \includegraphics[width=\textwidth]{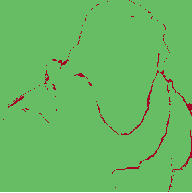}
        \includegraphics[width=\textwidth]{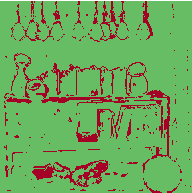}
        \includegraphics[width=\textwidth]{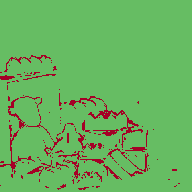}
        \subcaption{Ours BadPix}\label{fig:ours_badpix}
    \end{subfigure}
    \caption{
        \textbf{Results} on the HCI 4D Light Field Benchmark~\cite{Honauer2016} compared to the best learning-based competitor EPINET~\cite{Shin2018}.
        The BadPix score in \textbf{(e)} and \textbf{(f)} shows all pixels (red) exceeding an absolute distance of $0.07$ to the ground truth. 
    }
    \label{fig:results}
\end{figure*}
\begin{figure}[!htb]
    \includegraphics[width=\linewidth]{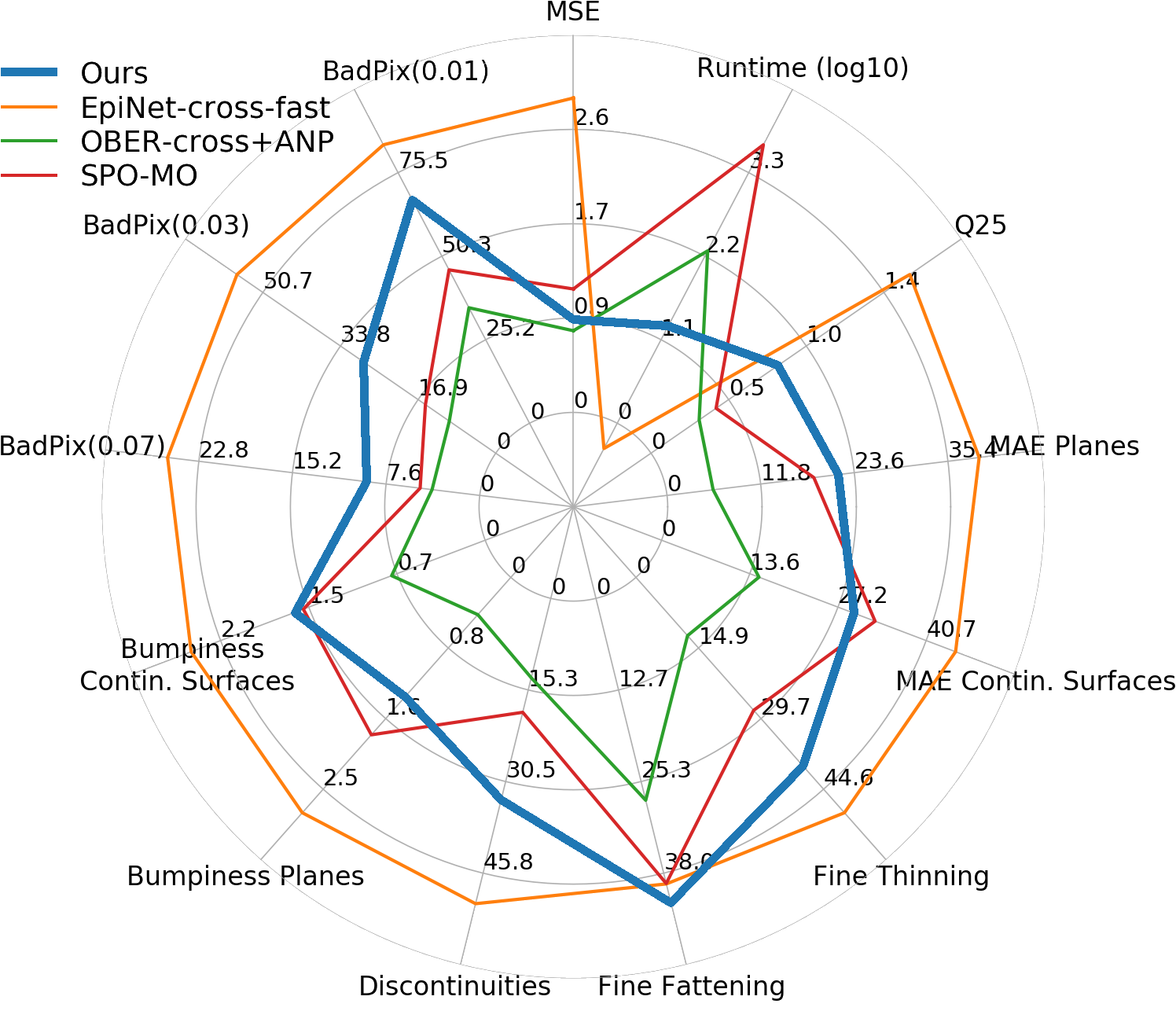}
    \caption{
        \textbf{Qualitative results on synthetic data.}
        We outperform EPINET-Cross~\cite{Shin2018} on $11$ out of $13$ metrics.
    }
    \label{fig:radar}
\end{figure}
For the quantitative evaluation in Figure~\ref{fig:radar}, we plot $13$ error measures from the HCI 4D Light Field Benchmark, for details please refer to \cite{Honauer2016}.
Our method outperforms EPINET-Cross~\cite{Shin2018}, in $11$ out of $13$ metrics with a close tie of the other two.
Because our network is run several times (once for each \ac{EPI}-Shift), the runtime increases linearly with the disparity range and is therefore slightly higher compared to~\cite{Shin2018}.
However, our approach is still significantly faster than the non-learning approaches.
Our work closes the performance gap to optimization-based methods while keeping all advantages of deep learning like fewer hyper parameters and learned instead of hand-crafted heuristics.
We evaluated our method on four photo-realistic scenes of the publicly available HCI 4D Light Field Benchmark~\cite{Honauer2016} that were not part of the training dataset.
Sadly, the benchmark uploading website is not functional anymore, hence no official submission was possible yet.
In Figure~\ref{fig:results} we show a qualitative comparison with EPINET~\cite{Shin2018}.
Note, that we use the cross setup for EPINET which uses the same $17$ views subset of the full light field that is used by our approach.
In addition to EPINET, the quantitative evaluation in Figure~\ref{fig:radar} also includes two state-of-the-art optimization based methods (OBER~\cite{Schilling2018} and SPO-MO \cite{Zhang2016}), ranking first and third in the official benchmark.

Our method provides considerably better quality at the disparity extremes (Scene 3 and 4, background) due to the improved generalization enabled by \ac{EPI}-Shift.
Also note the improved performance on non-textured surfaces (Scene 1, beige box) caused by the large receptive field of the U-Net which enables better smoothing and long-range reasoning.
Unfortunately, the U-Net also seems to be more prone to noise at object boundaries which are not disparity label boundaries (compare Scene 1, dark box), although similar artifacts can be observed with EPINET (compare Scene 3, books).

\subsection{Results on Real Recordings}
We also evaluated our method on images recorded by a cross light field setup with a disparity range of $[0, 12]$, consisting of $17$ cameras.
As the authors of~\cite{Shin2018} did not provide us with the pre-trained parameters for the cross-version of EPINET upon request, we trained EPINET-Cross based on their implementation for four days.
Figure~\ref{fig:teaser} (left) shows one of the results.
As expected, EPINET is only able to predict within the small training data disparity range of $[-3.5, 3.5]$, present in the background. 
In contrast, our~\ac{EPI}-Shift reconstructs the disparity in the whole range of $[0, 12]$.

%

\section{Conclusion}
To summarize, we have introduced a new learning-based approach for depth estimation from wide-baseline light field recordings. The key idea of our approach is to use so-called EPI-Shifts, similar to plane sweep volumes for stereo depth estimation. 
This approach improves the generalization capability of~\ac{CNN} based depth estimation and enables us to increase the receptive field using a U-Net which delivers better smoothing and reduces artifacts, thanks to long range reasoning.
Combining these two advantages leads to state-of-the-art performance, as demonstrated on a publicly available light field benchmark.
Furthermore, the EPI-Shift concept enables depth estimation with large baseline light fields, while the training data only exhibits small disparities. We also demonstrate results on a real world recording. 

\section{Acknowledgement}
We thank the Center for Information Services and High Performance Computing (ZIH) at TU Dresden for generous allocations of computer time.

{\small
\bibliographystyle{ieee}
\bibliography{refs}
}

\newpage
\appendix

\renewcommand\thefigure{\thesection.\arabic{figure}}
\setcounter{figure}{0}
\renewcommand\theequation{\thesection.\arabic{equation}}
\setcounter{equation}{0}
\renewcommand\thetable{\thesection.\arabic{table}}
\setcounter{table}{0}

\onecolumn

\renewcommand{\textfraction}{0.0}
\restylefloat{figure}
\section{Additional Results}

\begin{figure}[h]
    \centering
    \begin{subfigure}{0.2243\linewidth}
        \includegraphics[width=\textwidth]{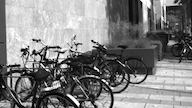}
        \includegraphics[width=\textwidth]{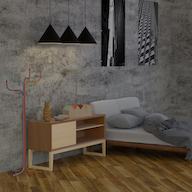}
        \includegraphics[width=\textwidth]{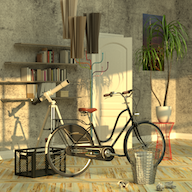}
        \includegraphics[width=\textwidth]{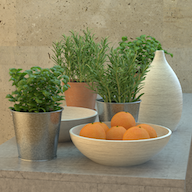}
        \includegraphics[width=\textwidth]{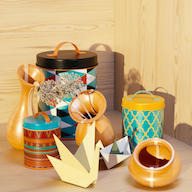}
        \subcaption{Center View}\label{fig:app_center}
    \end{subfigure}
    \begin{subfigure}{0.2243\linewidth}
        \includegraphics[width=\textwidth]{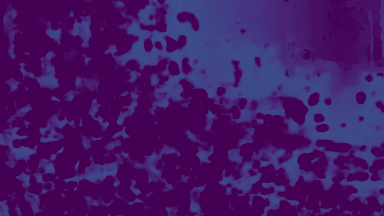}
        \includegraphics[width=\textwidth]{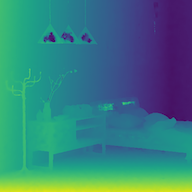}
        \includegraphics[width=\textwidth]{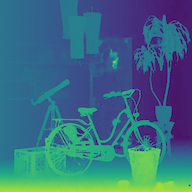}
        \includegraphics[width=\textwidth]{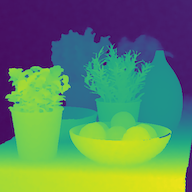}
        \includegraphics[width=\textwidth]{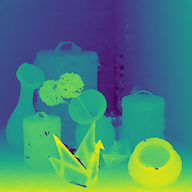}
        \subcaption{EPINET~\cite{Shin2018}}\label{fig:app_epinet}
    \end{subfigure}
    \begin{subfigure}{0.2243\linewidth}
        \includegraphics[width=\textwidth]{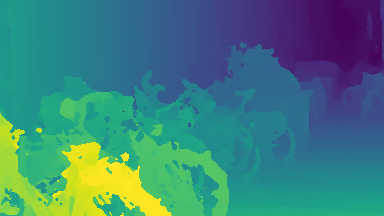}
        \includegraphics[width=\textwidth]{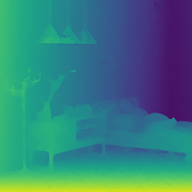}
        \includegraphics[width=\textwidth]{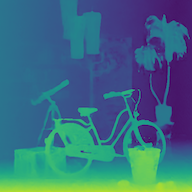}
        \includegraphics[width=\textwidth]{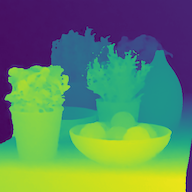}
        \includegraphics[width=\textwidth]{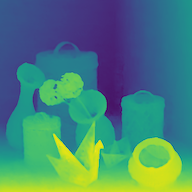}
        \subcaption{Ours}\label{fig:app_ours}
    \end{subfigure}
    \caption{
        \textbf{Results} on a second real scene (top) and four additional benchmark scenes~\cite{Honauer2016} without publicly available ground truth.
        The \textbf{real recording} (top) shows the limitation of EPINET~\cite{Shin2018} to the disparity range of the training data.
        \textbf{Additional benchmark scenes} show an improvement in non-textured areas and at extreme disparities (compare Scene 4 (the last scene), background) but also slightly more blurry results of our method.
        However, blur only occurs within the small regression intervals if two objects are part of the same depth label.
        At extreme disparities (compare Scene 3, background), our method also performs better due to the hard transitions between adjacent labels.
        \label{fig:results_test}
    }
\end{figure}

\begin{figure*}
    \centering
    \begin{subfigure}{0.162\textwidth}
        \includegraphics[width=\textwidth]{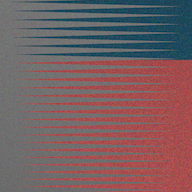}
        \includegraphics[width=\textwidth]{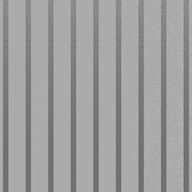}
        \includegraphics[width=\textwidth]{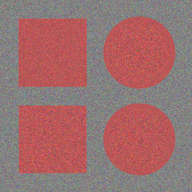}
        \includegraphics[width=\textwidth]{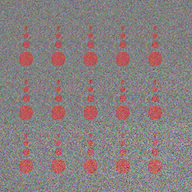}
        \subcaption{Center View}\label{fig:center}
    \end{subfigure}
    \begin{subfigure}{0.162\textwidth}
        \includegraphics[width=\textwidth]{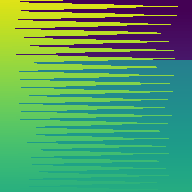}
        \includegraphics[width=\textwidth]{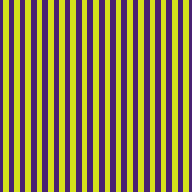}
        \includegraphics[width=\textwidth]{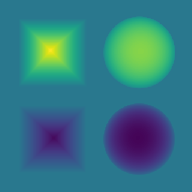}
        \includegraphics[width=\textwidth]{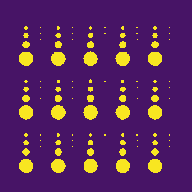}
        \subcaption{Ground Truth}\label{fig:gt}
    \end{subfigure}
    \begin{subfigure}{0.162\textwidth}
        \includegraphics[width=\textwidth]{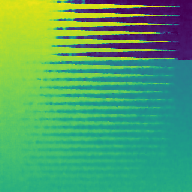}
        \includegraphics[width=\textwidth]{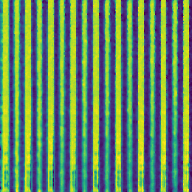}
        \includegraphics[width=\textwidth]{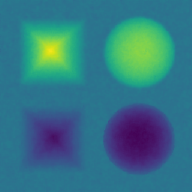}
        \includegraphics[width=\textwidth]{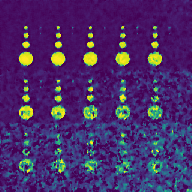}
        \subcaption{EPINET~\cite{Shin2018}}\label{fig:epinet}
    \end{subfigure}
    \begin{subfigure}{0.162\textwidth}
        \includegraphics[width=\textwidth]{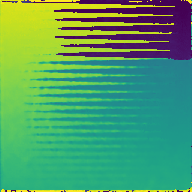}
        \includegraphics[width=\textwidth]{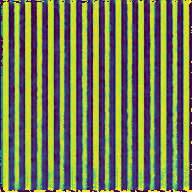}
        \includegraphics[width=\textwidth]{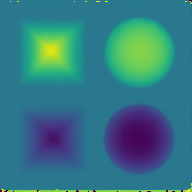}
        \includegraphics[width=\textwidth]{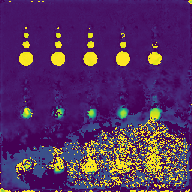}
        \subcaption{Ours}\label{fig:ours}
    \end{subfigure}
    \begin{subfigure}{0.162\textwidth}
        \includegraphics[width=\textwidth]{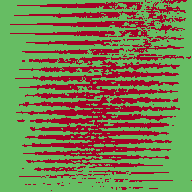}
        \includegraphics[width=\textwidth]{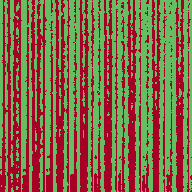}
        \includegraphics[width=\textwidth]{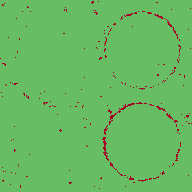}
        \includegraphics[width=\textwidth]{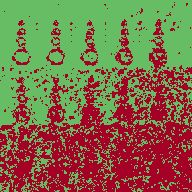}
        \subcaption{EPINET BadPix}\label{fig:epinet_badpix}
    \end{subfigure}
    \begin{subfigure}{0.162\textwidth}
        \includegraphics[width=\textwidth]{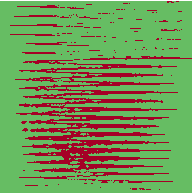}
        \includegraphics[width=\textwidth]{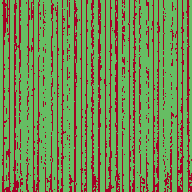}
        \includegraphics[width=\textwidth]{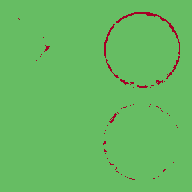}
        \includegraphics[width=\textwidth]{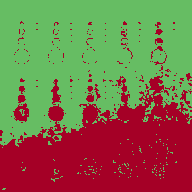}
        \subcaption{Ours BadPix}\label{fig:ours_badpix}
    \end{subfigure}
    \caption{
        \textbf{Results} on ``Stratified'' scenes of the HCI 4D Light Field Benchmark~\cite{Honauer2016}.
        The BadPix score in \textbf{(e)} and \textbf{(f)} shows all pixels (red) exceeding an $\mathcal{L}^1$-distance of $0.07$ to the ground truth. 
        Note the~\textbf{improved smoothness} on flat surfaces due to the U-Net architecture (compare Scene 3, background).
        Also note the \textbf{failure case} of our method in Scene 4, caused by strong noise occuring only in the bottom of the image.
        In those cases,~\ac{EPI}-Shift causes misclassifications, leading to stronger artifacts than EPINET~\cite{Shin2018}.
        \label{fig:results_stratified}
    }
\end{figure*}

\end{document}